\documentclass{article}

\usepackage[preprint]{neurips_2025}

\usepackage[utf8]{inputenc} 
\usepackage[T1]{fontenc}    
\usepackage{hyperref}       
\usepackage{url}            
\usepackage{booktabs}       
\usepackage{amsfonts}       
\usepackage{nicefrac}       
\usepackage{microtype}      
\usepackage{xcolor}         

\usepackage{amsfonts,amsmath,amssymb,amsthm,mathtools}
\usepackage{natbib}
\usepackage{booktabs}
\usepackage{wrapfig}
\usepackage{subcaption}
\usepackage{listings}
\usepackage{xcolor}

\definecolor{codegreen}{rgb}{0,0.6,0}
\definecolor{codegray}{rgb}{0.5,0.5,0.5}
\definecolor{codepurple}{rgb}{0.58,0,0.82}
\definecolor{backcolour}{rgb}{0.95,0.95,0.92}

\lstdefinestyle{mystyle}{
    backgroundcolor=\color{backcolour},   
    commentstyle=\color{codegreen},
    keywordstyle=\color{magenta},
    numberstyle=\tiny\color{codegray},
    stringstyle=\color{codepurple},
    basicstyle=\ttfamily\footnotesize,
    breakatwhitespace=false,         
    breaklines=true,                 
    captionpos=b,                    
    keepspaces=true,                 
    numbers=left,                    
    numbersep=5pt,                  
    showspaces=false,                
    showstringspaces=false,
    showtabs=false,                  
    tabsize=2
}

\title{Rodent-Bench}
\title{
\begin{minipage}{0.12\textwidth}
  \includegraphics[width=1.2\linewidth]{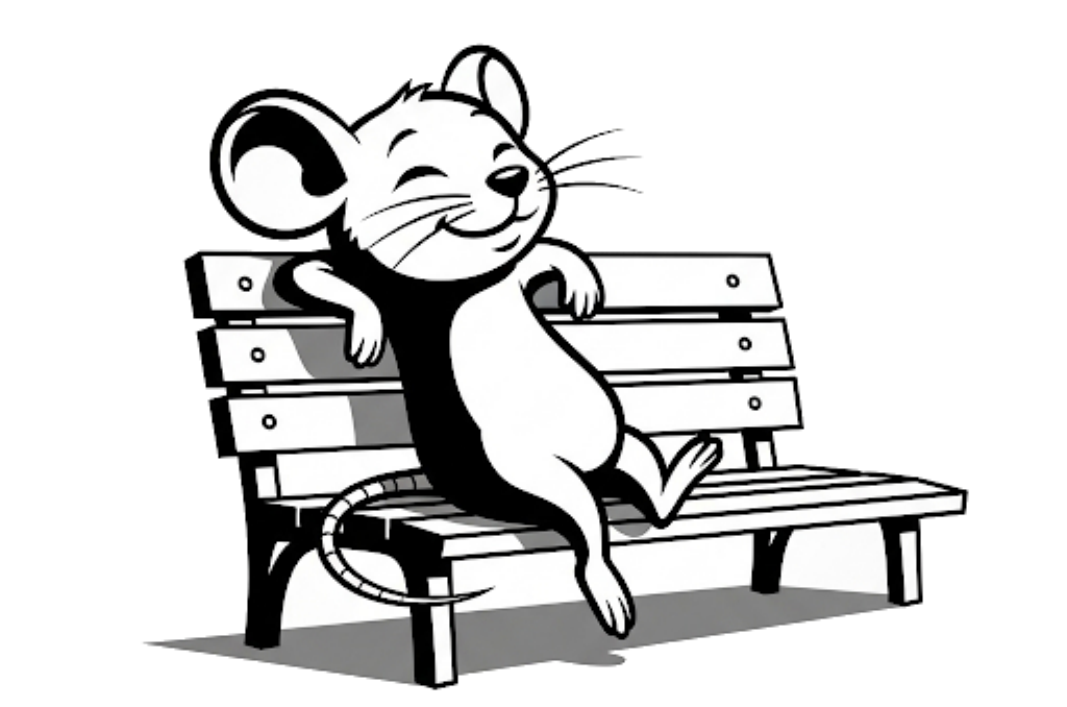}
\end{minipage}
\begin{minipage}{0.85\textwidth}
\centering
\hspace{-2em}
\textbf{Rodent-Bench}
\end{minipage}
}

\author{
    Thomas Heap \\
    University of Bristol \\
    Bristol, UK \\
    \texttt{thomas.heap@bristol.ac.uk} \\
    \And
    Adriana Casado Rodriguez \\
    University of Bristol \\
    Bristol, UK \\
    \And
    Emma N. Cahill \\
    University of Bristol \\
    Bristol, UK \\
    \texttt{ORCID (0000-0003-3054-1708)}
    \And
    Laurence Aitchison \\
    University of Bristol \\
    Bristol, UK \\
}

\begin{document}

\maketitle

\begin{abstract}
    We present Rodent-Bench, a novel benchmark designed to evaluate the ability of Multimodal Large Language Models (MLLMs) to annotate rodent behaviour footage. We evaluate state-of-the-art MLLMs, including Gemini-2.5-Pro, Gemini-2.5-Flash and Qwen-VL-Max, using this benchmark and find that none of these models perform strongly enough to be used as an assistant for this task. Our benchmark encompasses diverse datasets spanning multiple behavioral paradigms including social interactions, grooming, scratching, and freezing behaviors, with videos ranging from 10 minutes to 35 minutes in length. We provide two benchmark versions to accommodate varying model capabilities and establish standardized evaluation metrics including second-wise accuracy, macro F1, mean average precision, mutual information, and Matthew's correlation coefficient. While some models show modest performance on certain datasets (notably grooming detection), overall results reveal significant challenges in temporal segmentation, handling extended video sequences, and distinguishing subtle behavioral states. Our analysis identifies key limitations in current MLLMs for scientific video annotation and provides insights for future model development. Rodent-Bench serves as a foundation for tracking progress toward reliable automated behavioral annotation in neuroscience research.
\end{abstract}

\section{Introduction}

Behavioral analysis is fundamental to neuroscience and biomedical research, yet manual annotation of animal behavior videos remains a time-consuming bottleneck that limits research scale and reproducibility \citep{sturm2020deep, mathis2020deep}. While Multimodal Large Language Models (MLLMs) have shown impressive capabilities in vision-language tasks \citep{fu2024video, yin2024survey}, their application to specialized scientific domains like behavioral analysis remains largely unexplored. MLLMs offer particular promise for scientific annotation tasks as they can potentially handle diverse behavioral paradigms through natural language instructions without requiring specialized model training for each new behavior or experimental setup.

Unlike general computer vision tasks, behavioral analysis requires models to identify subtle, context-dependent actions, maintain temporal coherence across extended sequences, and produce structured outputs aligned with ethological frameworks. Traditional computer vision approaches require training specialized models for each behavioral task, but MLLMs could streamline this process by accepting task descriptions in natural language and adapting to new behaviors without retraining. Existing video understanding benchmarks inadequately address these scientific requirements, creating a significant gap between current MLLM capabilities and practical research applications.

We present \textbf{Rodent-Bench-Short} and \textbf{Rodent-Bench-Long}, the first comprehensive benchmarks for evaluating MLLMs on rodent behavioral annotation tasks. Our benchmarks encompasse diverse datasets spanning multiple behavioral paradigms and provides standardized evaluation metrics to assess current model capabilities. We evaluate state-of-the-art MLLMs including Gemini-2.5-Pro, Gemini-2.5-Flash, and Qwen-VL-Max, revealing significant performance gaps that limit their utility as research assistants. While some of these models show fair performance on some datasets, our analysis identifies specific challenges in temporal segmentation, long video processing, and handling varied experimental conditions, providing insights for future improvements in scientific applications of multimodal models.

\section{Related Work}

The emergence of Multimodal Large Language Models (MLLMs) has opened new possibilities for video understanding tasks across diverse domains. Recent comprehensive benchmarks such as Video-MME \citep{fu2024video} have evaluated state-of-the-art MLLMs including GPT-4 and Gemini on video analysis tasks, revealing significant challenges in temporal reasoning and long-form video understanding. Surveys on video understanding with large language models \citep{tang2025video, wang2024seconds} highlight the emergent capabilities of these systems for multi-granularity reasoning, while identifying key limitations in handling long-form videos and maintaining alignment between visual and textual modalities. Despite these advances, the application of MLLMs to specialized scientific domains remains under-explored, with recent work suggesting significant potential for leveraging these models in natural science research \citep{yin2024survey, testard2024data}.

Traditional animal behavior analysis has undergone significant transformation with the advent of deep learning and computer vision techniques. Deep learning-based behavioral analysis systems have demonstrated the ability to reach human-level accuracy in recognizing specific ethological behaviors \citep{sturm2020deep}, with markerless pose estimation tools like DeepLabCut enabling robust tracking of individual body parts in freely moving rodents \citep{mathis2020deep}. Specialized tools such as DeepBehavior \citep{arac2019deep}, ezTrack \citep{pennington2019eztrack}, MoSeq \citep{wiltschko2015mapping}, SLEAP \cite{pereira2022sleap} and real-time behavior recognition systems \citep{dechaumont2022real} have been developed specifically for automated analysis of animal behavior. However, these systems typically require task-specific training and lack the flexibility and generalization capabilities that modern MLLMs potentially offer. The specific application of MLLMs to behavioral annotation tasks in laboratory settings remains largely unexplored, representing a significant gap that our Rodent-Bench benchmark aims to address.

\section{Rodent-Bench}

We produced two benchmarks: \textbf{Rodent-Bench-Short}, with videos up to 10 minutes long; and \textbf{Rodent-Bench-Long}, with videos up to 35 minutes long. We created these two versions because current MLLMs have varying video length limitations—while models like Gemini can process videos up to 1 hour, others like Qwen-VL-Max are restricted to 10 minutes or less. This dual-benchmark approach ensures compatibility across all evaluated models and enables investigation of how video length affects annotation performance.

Along with this we suggest evaluation metrics. The task posed to the MLLM is to annotate each video, determining which of a fixed set of behaviours is occurring and to produce a JSON file segmenting each video into discrete non-overlapping time segments with behaviour labels.

\begin{figure}[!htb]
    \centering
    \includegraphics[width=0.9\linewidth]{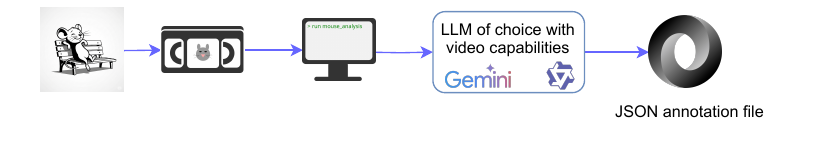}
    \caption{Workflow for annotating rodent videos.}
    \label{fig:workflow}
\end{figure}

\subsection{Data Collection}

We collected our data from several openly available datasets, as well as one private dataset which we now make freely available.

\textbf{Caltech Rodent Social Interactions (CalMS21):}

The CalMS21 dataset \citep{calms21} is intended for multi-agent behaviour modelling. It consists of footage of multiple rodents interacting socially, with 6 million un-labelled frames and 1 million labelled frames. The labelled frames consist of both frame level behaviour and pose tracking annotations. For our purposes we use the first 25 labelled videos in the training set. 

\textbf{Rodent Grooming Detection Annotated Dataset:}

The rodent grooming dataset \citep{grooming} was collected in order to train a neural network rodent grooming classifier. It consists of 1,253 video clips with 2,637,363 frames. Each frame is labelled ``Grooming'' or ``Not Grooming''. We use the first 25 videos in the training set for our eval.

\textbf{Mouse-Ventral 1\&2:}
We use the Mouse-Ventral 1\&2 subsets of the Deep Ethogram dataset \citep{deepethogram}. These consists of 30 minute videos of a rodents shot from below, 16 for MV2 and 28 for MV1, the videos are annotated with behaviour labels. In the Mouse-Ventral1 subset the rodents are either ``grooming'', ``digging'', ``scratching'', ``licking'' or ``background'' (neither scratching nor licking). In the Mouse-Ventral2 subset the rodents are either ``scratching'', ``licking'' or ``background''. 

\textbf{Scratch-AID:}
The Scratch-AID dataset \citep{scratchaid} was collected to train a neural network CRNN rodent scratching classification model. The dataset consists of 40 videos of rodents shot from below, the rodents were injected with an itching agent causing them to scratch compulsively. The model trained especially for this task achieved 97.6\% recall and 96.9\% precision on previously unseen test videos.
\begin{figure}
    \centering
    \includegraphics[width=\linewidth]{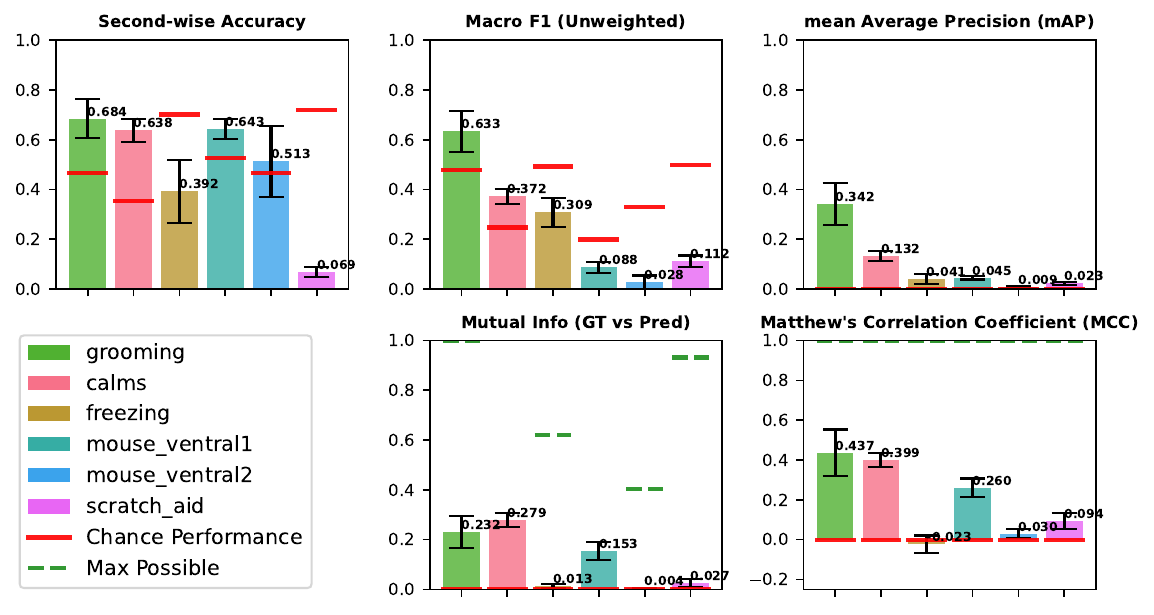}
    \caption{Performance metrics for Gemini-2.5-Pro across all datasets. Each metric shows substantial variation across behavioral paradigms, with the grooming detection dataset achieving the highest performance across most metrics. Social behaviors (CalMS21) show moderate performance, while challenging datasets like freezing and scratch detection exhibit poor performance approaching chance levels. Dashed lines indicate theoretical maximum performance where applicable. Error bars represent $2 \times$ standard error across videos within each dataset. The consistently low performance on certain datasets highlights the difficulty of fine-grained temporal behavioral annotation for current MLLMs.}
    \label{fig:pro_comparison}
\end{figure}

\textbf{Freezing:}
Our collaborators have given us access to nine videos of rodents displaying a ``freezing'' behaviour. This is a behavior distinct from resting, and is characterised by the ears being oriented towards the front indicating alert but immobile behavior \citep{blanchard1969passive}. There are three types of videos, and three videos of each type: Low freezing, high freezing and extinction. Extinction is a behavioral paradigm where the conditioned freezing response is gradually reduced through repeated exposure to the conditioned stimulus without the unconditioned stimulus.

This dataset is particularly important for evaluating MLLMs because freezing behavior presents challenges that traditional pose estimation approaches cannot address. While tools like DeepLabCut excel at tracking body parts and movements, they cannot distinguish between freezing (an active fear response) and other motionless states such as sleeping, resting, or general inactivity. These behaviors are not easily distinguishable when relying solely on pose or movement data. MLLMs, with their ability to integrate visual context, temporal patterns, and behavioral understanding, may offer advantages for this subtle but scientifically important distinction.

\textbf{Rodent-Bench-Short:}
Some MLLMs will not accept long video files (30 minutes to an hour), so to evaluate these models we produce a shortened version of the dataset in which any file longer than 10 minutes is shortened to that length. We evaluate all models on both datasets for comparison.

\subsection{Metrics}

\textbf{Second-wise accuracy:} We treat each second as a binary classification problem: is the behaviour in that second correctly classified or not. We then report the proportion of seconds in which the behaviour was correctly classified. 

\textbf{Macro F1:} We calculate the F1 score for each class and average with no weighting.

For each behavior class $c$:
\begin{align}
\text{Precision}_c &= \frac{TP_c}{TP_c + FP_c} \\
\text{Recall}_c &= \frac{TP_c}{TP_c + FN_c} \\
\text{F1}_c &= \frac{2 \cdot \text{Precision}_c \cdot \text{Recall}_c}{\text{Precision}_c + \text{Recall}_c}
\end{align}

Macro F1 is the unweighted average across all classes:
\begin{align}
\text{Macro F1} = \frac{1}{|C|} \sum_{c \in C} \text{F1}_c
\end{align}

where $|C|$ is the number of behavior classes, $TP_c$ is true positives for class $c$, $FP_c$ is false positives, and $FN_c$ is false negatives.

\textbf{mean Average Precision (mAP):} This is calculated by comparing predicted and ground truth behaviour segments across a range of IoU (Intersection over Union) thresholds (from 0.1 to 0.9) \citep{henderson2016end}. For each threshold, we match predicted segments to ground truth segments of the same behaviour if their IoU exceeds the threshold, counting true positives (TP), true negatives (TN), false positives (FP), and false negatives (FN). Precision and recall are computed at each threshold, and the average precision is accumulated as the sum of precision values weighted by the change in recall, this is to approximate the area under the precision-recall curve. The final mAP is the total of these values, providing a single metric that summarizes how well the predictions align with the ground truth across different levels of overlap.

\textbf{Mutual Information: } We calculate the mutual information between the ground-truth second-wise labels and the predicted labels.

\textbf{Matthew's Correlation Coefficient (MCC):} This is a correlation coefficient between -1 and 1. It is calculated:

\begin{align*}
\mathrm{MCC} = \frac{ (TP \times TN) - (FP \times FN) }
{ \sqrt{ (TP + FP)(TP + FN)(TN + FP)(TN + FN) } }
\end{align*}

\textbf{Dataset label entropy weighted Matthew's Correlation Coefficient:} To provide a singular score which takes into account differing datasets ``difficulty'':

\[
w_i = \frac{(H_i + \epsilon) \cdot T_i}{\sum_j (H_j + \epsilon) \cdot T_j}
\]

Where, $H_i$ is the entropy of dataset $i$, $T_i$ is the duration of dataset $i$ in seconds, $\epsilon = 10^{-8}$ is a small constant to avoid zero weights. This weighting scheme assigns higher weight to longer datasets with more diverse labels, which should be more challenging.

We prioritize mutual information, MCC, and mAP metrics for our primary analysis. Mutual information and MCC provide interpretable baselines, both equalling zero when predictions are statistically independent of ground truth labels, making chance-level performance easily identifiable across all datasets. The mAP metric is valuable for evaluating temporal segmentation quality, as it directly measures how well predicted behavioral segments align with ground truth boundaries across multiple IoU thresholds. In contrast, metrics like second-wise accuracy and macro F1 have dataset-dependent chance baselines that vary with class distributions and choice of random baseline strategy (uniform random vs. frequency-matched random prediction), complicating cross-dataset comparisons and performance interpretation.

\section{Experiments}

To provide an idea of how models currently perform we evaluate some of the available MLLM's on this benchmark.

\subsection{Experimental Setup}
\textbf{Models} We evaluate our benchmark on three MLLMs. As of July 2025, a number of MLLMs which claim support for video actually just sample frames from the video at regular intervals and use these (e.g. Qwen-VL-Max). We use Gemini-2.5-Flash, Gemini-2.5-Pro and Qwen-VL-Max. Specifications of these models can be found in Appendix \ref{app:models}.

\begin{figure}[t]
  \centering
  \begin{subfigure}{0.45\textwidth}
    \centering
    \includegraphics[width=\textwidth]{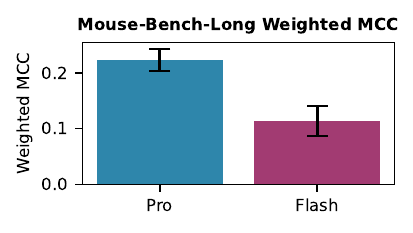}
    \caption{}
    \label{fig:weighted_mcc}
  \end{subfigure}
  \hfill
  \begin{subfigure}{0.45\textwidth}
    \centering
    \includegraphics[width=\textwidth]{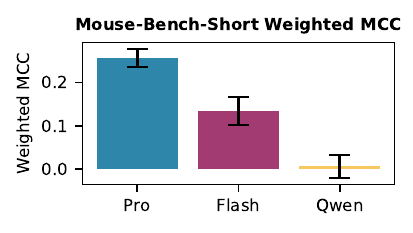}
    \caption{}
    \label{fig:weighted_mcc_short}
  \end{subfigure}
  \caption{Weighted Matthew's Correlation Coefficient (MCC) performance across models. (a) Rodent-Bench-Long: Gemini-2.5-Pro achieves the highest performance with lower variance compared to Gemini-2.5-Flash. (b) Rodent-Bench-Short: Similar performance hierarchy with Gemini-2.5-Pro outperforming Flash, while Qwen-VL-Max shows near-chance performance. Error bars represent $2 \times$ standard error across datasets. All models show modest performance levels, indicating substantial room for improvement in behavioral annotation tasks.}
\end{figure}

\textbf{Prompting Strategy} Because our dataset is heterogenous we use different prompts for each of the sub-datasets, these appear in appendix \ref{app:prompts}. Despite this we tried to use a similar prompt for each dataset in the interest of fairness. The same prompt is used for all models.

\textbf{Computational Cost:} For the complete benchmark, costs are approximately \$7 for Gemini-2.5-Pro, \$7 for Qwen-VL-Max, and under \$1 for Gemini-2.5-Flash.

\subsection{Results}

We evaluate three MLLMs on our benchmark. Further figures showing individual dataset performance for each model can be found in appendices \ref{app:pro}, \ref{app:flash} and \ref{app:qwen}.

In figure \ref{fig:weighted_mcc} you can see that Gemini Pro outperforms Flash in both absolute performance, as well as variability. 

In \ref{fig:weighted_mcc_short}, evaluating on the shortened dataset, we see a similar comparison between Pro and Flash, while Qwen-VL-Max performs no better than chance.

We notice that both Flash and Qwen models struggle with correct formatting, sometimes the wrong key for certain segments (``end\_long\_time'' instead of ``end\_time'') or in the case of Qwen-VL it will simply stop partway through a segment, rendering the JSON file unreadable (without modifications).

We speculate that these models perform strongest on datasets with shorter videos on average, that have clearly defined labels which depend only on behaviour (i.e do not require the rodent to be in a particular position in the cage when acting for that label to apply), and have behaviours that last at least a few seconds. They perform weakly on videos which have visual filters applied but that require some degree of colour recognition for labelling (i.e ``the feeding box is at the back of the cage and is black''), or that are taken from a ``non-standard'' (i.e not front facing, from above or from below) camera angle, or that have very short or ambiguous behaviours. For instance the freezing dataset features behaviours shorter than a second, and the difference between the rodent ``freezing'' and simply not moving is quite subtle.

\section{Limitations}
Our benchmark has several important limitations. First, ground-truth annotations were taken from existing datasets with varying labelling schemes and quality standards, potentially containing inconsistencies that affect evaluation reliability. Second, we lack human annotator baselines to contextualize model performance—while current MLLMs perform poorly, we cannot determine how their accuracy compares to average human annotators on these specific videos.

Third, our evaluation uses zero-shot inference without fine-tuning or extensive prompt optimization. While this ensures fair comparison across models, it may underestimate achievable performance through model adaptation or specialized prompting strategies. Additionally, dataset-specific prompts introduce variability that could advantage certain models.

Finally, our evaluation is limited to three commercially available models with native video processing, and the rapid pace of model development means newer capabilities may alter these findings. Despite these limitations, Rodent-Bench provides a valuable initial assessment of current MLLM capabilities for scientific behavioral annotation tasks.

\section{Conclusion}

We introduced Rodent-Bench, the first comprehensive benchmark for evaluating Multimodal Large Language Models on scientific behavioral annotation tasks. Our evaluation of state-of-the-art MLLMs—Gemini-2.5-Pro, Gemini-2.5-Flash, and Qwen-VL-Max—reveals that current models perform substantially below the accuracy levels required for practical deployment as research assistants in behavioral neuroscience.

While MLLMs showed modest success on certain datasets (notably grooming detection), performance varied dramatically across behavioral paradigms. Models struggled particularly with subtle temporal distinctions, brief behavioral episodes, and tasks requiring integration of spatial context with behavioral understanding. The freezing behavior dataset exemplified these challenges, where distinguishing between active freezing responses and passive inactivity proved difficult even for advanced multimodal systems.

Our findings highlight several critical areas for improvement. First, enhanced temporal reasoning capabilities are needed to handle the fine-grained segmentation required for behavioral analysis. Second, models must develop better contextual understanding to distinguish between visually similar but behaviorally distinct states. Finally, output formatting consistency remains a practical barrier, with some models frequently producing malformed JSON responses that complicate automated processing.

Despite current limitations, Rodent-Bench establishes a foundation for tracking progress in scientific applications of multimodal AI. The benchmark's diverse behavioral paradigms and standardized evaluation framework provide a testbed for future model improvements. As MLLMs advance, their potential to democratize behavioral annotation, eliminating the need for specialized model training for each experimental paradigm, remains promising. Rodent-Bench will enable researchers to objectively assess when these models achieve the reliability threshold necessary for practical scientific deployment.

The gap between current capabilities and scientific requirements underscores the need for continued research at the intersection of multimodal AI and domain-specific applications. Our benchmark contributes to this effort by providing concrete evaluation targets and highlighting the unique challenges that scientific video understanding presents to current generation models.

\bibliography{refs}
\bibliographystyle{icml2024}

\newpage
\appendix

\section{Implementation Details}
\label{app:implementation}

\subsection{Model Access and Configuration}

\textbf{Gemini Models:} We access Gemini-2.5-Pro and Gemini-2.5-Flash through Google's Vertex AI GenAI SDK. Videos are processed directly from Google Cloud Storage URIs using the native video input capabilities. The response format is constrained to JSON using structured output schemas specific to each dataset's behavior categories.

\textbf{Qwen-VL-Max:} We access Qwen-VL-Max through Alibaba's DashScope API using the OpenAI-compatible interface. 

\subsection{Video Processing Pipeline}

Each video is processed independently with dataset-specific prompts that include behavior definitions, temporal annotation requirements, and output format specifications. 

\textbf{Structured Output Schemas:} We define Pydantic models for each dataset's behavior categories to ensure consistent JSON output formatting. The following is the model for CaLMS21:

\begin{lstlisting}[language=Python]
class RodentBehaviorSegment(BaseModel):
    segment_number: int = Field(..., description="Segment number in order")
    start_time: str = Field(..., description="Start time in MM:SS format")
    end_time: str = Field(..., description="End time in MM:SS format")
    behavior: str = Field(..., description="Behavior label (e.g., attack, investigation, mount, other)")
\end{lstlisting}
\subsection{Batch Processing Implementation}

For Gemini models, we implement both individual and batch processing modes. Batch mode generates JSONL files conforming to Gemini's batch API requirements, uploads input files to Google Cloud Storage, and monitors job completion through the batch API. This approach significantly reduces API costs for large-scale evaluations while maintaining identical model configurations.

\textbf{Error Handling:} The system logs all API responses, including malformed outputs, to facilitate debugging. For models producing incomplete JSON (particularly Qwen-VL-Max), we save raw responses to text files for manual inspection. Output validation ensures all required fields are present and temporal segments are non-overlapping.

\newpage

\section{Model Specifications}
\label{app:models}

We evaluate three state-of-the-art multimodal large language models with native video understanding capabilities.

\subsection{Gemini-2.5-Pro}

\textbf{Key Specifications:}
\begin{itemize}
\item Maximum video length: 1 hour (without audio), 45 minutes (with audio)
\item Context window: 1,048,576 tokens (input), Maximum 65,535 tokens (output)
\item Maximum video file size: 2 GB
\end{itemize}

\subsection{Gemini-2.5-Flash}

\textbf{Key Specifications:}
\begin{itemize}
\item Maximum video length: 1 hour (without audio), 45 minutes (with audio)
\item Context window: 1,048,576 tokens (input), Maximum 65,535 tokens (output)
\item Maximum video file size: 2 GB
\end{itemize}

\subsection{Qwen-VL-Max}

Qwen-VL-Max is Alibaba Cloud's most advanced vision-language model. Unlike the Gemini models which process video natively, Qwen-VL extracts frames from video files for analysis, extracting one frame every 0.5 seconds when using the OpenAI SDK.

\textbf{Key Specifications:}
\begin{itemize}
\item Maximum video length: 10 minutes (Qwen2.5-VL series)
\item Context window: 129,024 input tokens, 8,192 output tokens
\item Maximum video file size: 1 GB (via URL), 10 MB (Base64 encoded)
\item Video processing: Frame extraction (no audio support)
\end{itemize}

\subsection{Model Selection Rationale}

We selected these models based on three criteria: (1) native video processing capabilities, (2) availability through stable APIs for reproducible evaluation, and (3) demonstrated performance on complex reasoning tasks. 

\newpage
\section{Datasets}
\label{app:datasets}

We include screenshots from each dataset, demonstrating each behavior. We additionally include a chart showing the proportion of each behavior in each dataset.

\begin{figure}[!h]
    \centering
    \includegraphics[width=\linewidth]{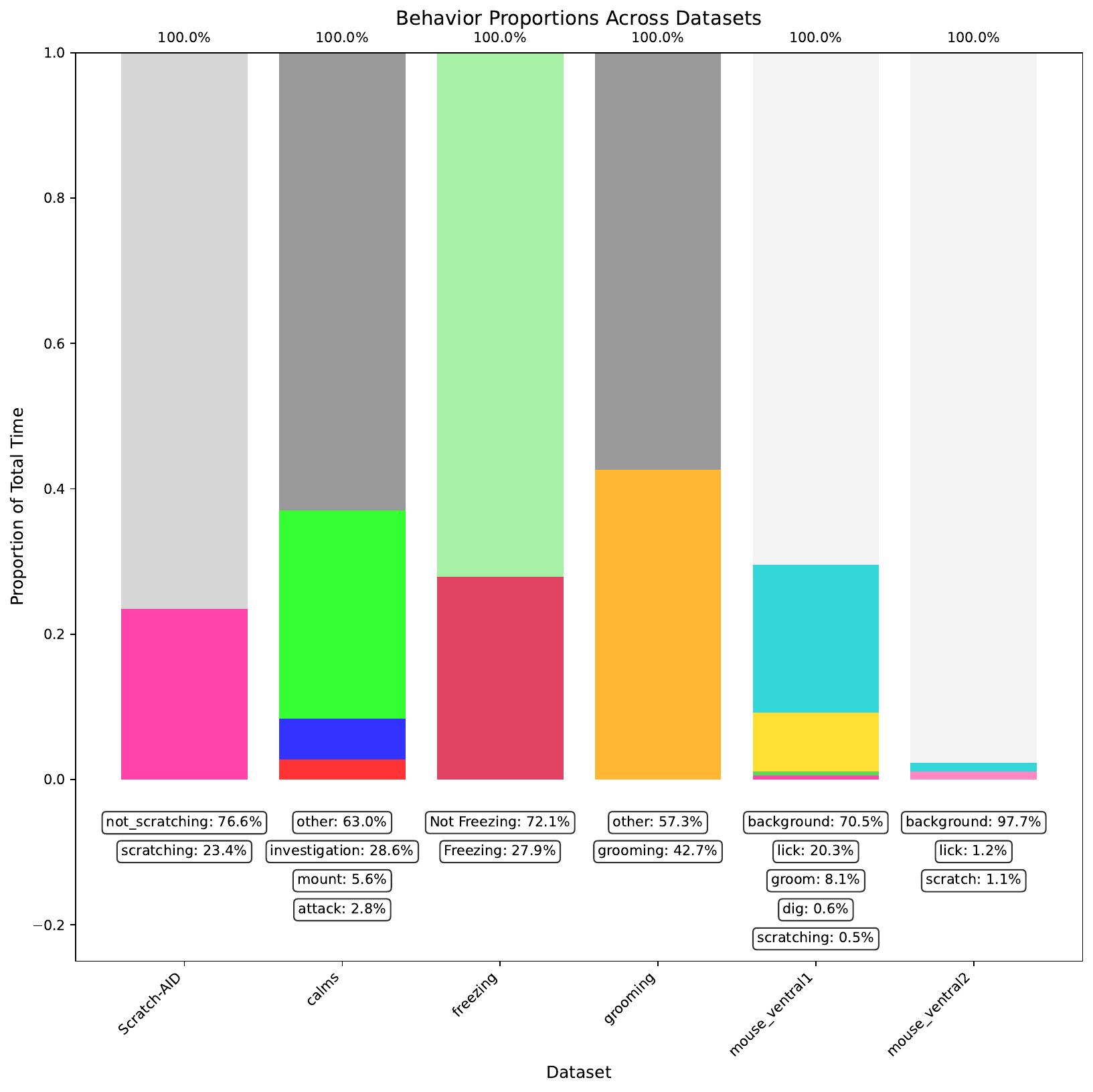}
    \caption{Behavior Proportions for each dataset.}
    \label{fig:proportions}
\end{figure}

\newpage

\subsection{Rodent-Bench}
\begin{table}[htbp]
\centering
\caption{Timing Statistics by Dataset}
\begin{tabular}{lcccc}
\hline
\textbf{Dataset} & \textbf{Average Time} & \textbf{Minimum Time} & \textbf{Maximum Time} & \textbf{Total Time} \\
                  & \textbf{(Mins)}       & \textbf{(Mins)}       & \textbf{(Mins)}       & \textbf{(Mins)}     \\
\hline
CalMS21          & 4.41                  & 1.02                  & 11.87                 & 110.22              \\
Freezing         & 14.35                 & 4.01                  & 32.67                 & 129.17              \\
Grooming         & 1.32                  & 0.34                  & 5.99                  & 32.88               \\
Rodent Ventral 1  & 8.33                  & 8.28                  & 8.33                  & 233.25              \\
Rodent Ventral 2  & 29.97                 & 29.97                 & 29.97                 & 479.57              \\
Scratch-AID      & 20.00                 & 20.00                 & 20.00                 & 300.03              \\
\hline
\end{tabular}
\label{tab:timing_stats}
\end{table}

\subsection{Rodent-Bench Short}

\begin{table}[htbp]
\centering
\caption{Timing Statistics by Dataset}
\begin{tabular}{lcccc}
\hline
\textbf{Dataset} & \textbf{Average Time} & \textbf{Minimum Time} & \textbf{Maximum Time} & \textbf{Total Time} \\
                  & \textbf{(Mins)}       & \textbf{(Mins)}       & \textbf{(Mins)}       & \textbf{(Mins)}     \\
\hline
CalMS21          & 4.30                  & 1.02                  & 9.98                  & 107.57              \\
Freezing         & 9.04                  & 4.01                  & 9.98                  & 81.33               \\
Grooming         & 2.87                  & 0.44                  & 9.98                  & 71.68               \\
Rodent Ventral 1  & 6.57                  & 0.34                  & 8.33                  & 183.89              \\
Rodent Ventral 2  & 9.26                  & 8.33                  & 9.98                  & 148.19              \\
Scratch-AID      & 9.98                  & 9.98                  & 9.99                  & 149.77              \\
\hline
\end{tabular}
\label{tab:timing_stats_short}
\end{table}

\newpage
\subsection{CaLMS21}
\begin{figure}[!h]
    \centering
    \includegraphics[width=0.95\linewidth]{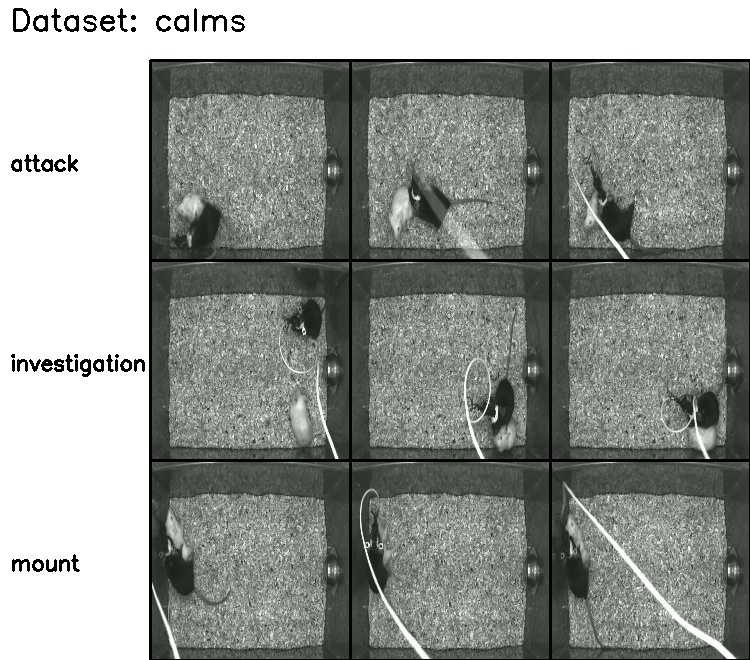}
    \caption{CaLMS21 Behaviors}
    \label{fig:calms_behaviors}
\end{figure}

\subsection{Rodent Grooming}
\begin{figure}[!h]
    \centering
    \includegraphics[width=0.95\linewidth]{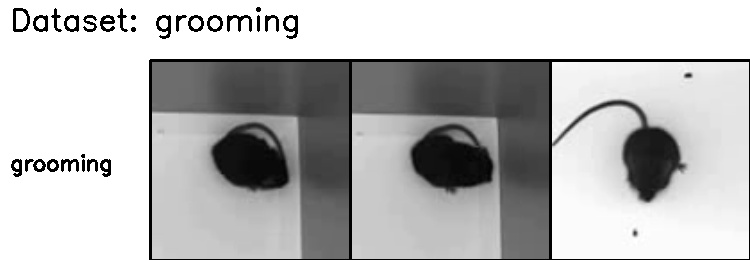}
    \caption{Rodent Grooming Behaviors}
    \label{fig:grooming_behaviors}
\end{figure}

\newpage
\subsection{Mouse-Ventral 1\&2}
\begin{figure}[!h]
    \centering
    \includegraphics[width=0.95\linewidth]{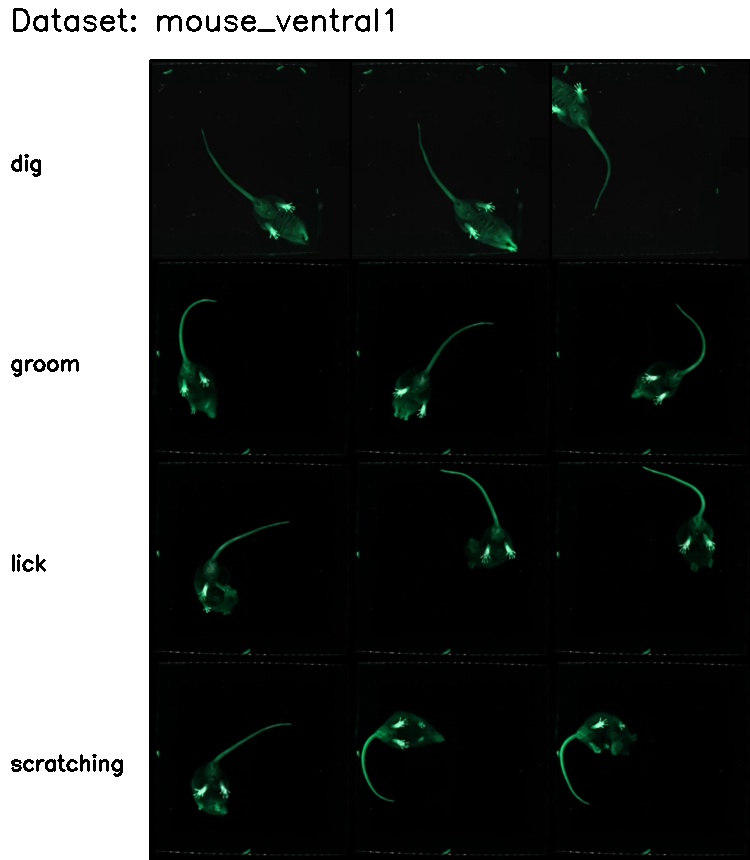}
    \caption{Mouse-Ventral 1 Behaviors}
    \label{fig:ventral1_behaviors}
\end{figure}

\begin{figure}[!h]
    \centering
    \includegraphics[width=0.95\linewidth]{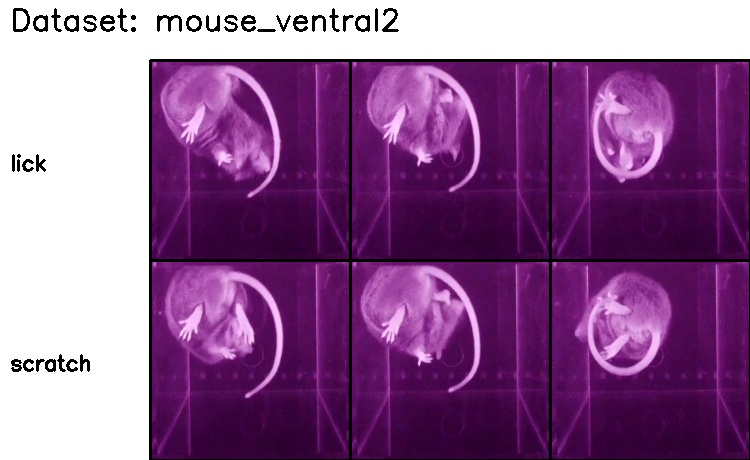}
    \caption{Mouse-Ventral 2 Behaviors}
    \label{fig:ventral2_behaviors}
\end{figure}

\newpage
\subsection{Scratch-AID}
\begin{figure}[!h]
    \centering
    \includegraphics[width=0.95\linewidth]{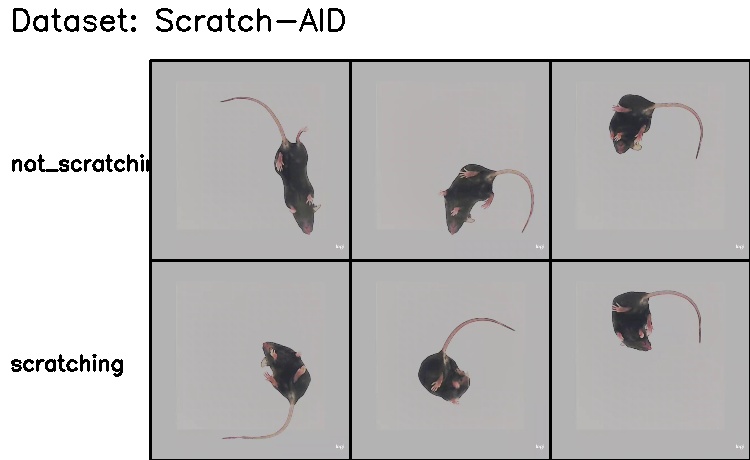}
    \caption{Scratch-AID Behaviors}
    \label{fig:scratch-aid_behaviors}
\end{figure}

\subsection{Freezing}
\begin{figure}[!h]
    \centering
    \includegraphics[width=0.95\linewidth]{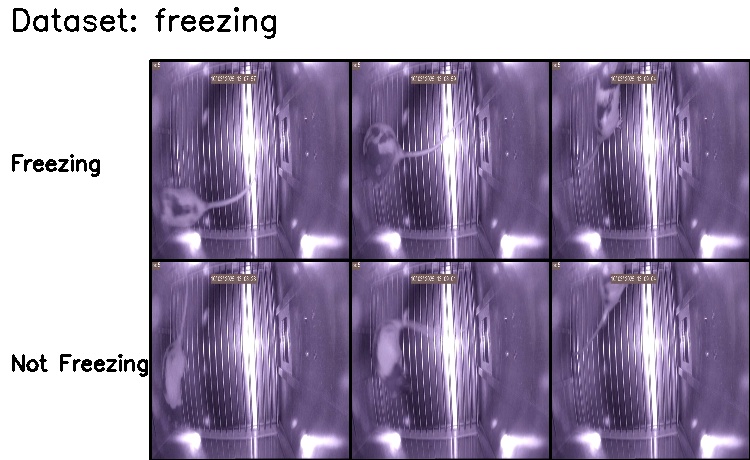}
    \caption{Freezing Behaviors}
    \label{fig:freezing_behaviors}
\end{figure}

\newpage
\section{Gemini-Pro Results}
\label{app:pro}

\subsection{Rodent-Bench}

\begin{figure}[!h]
    \centering
    \includegraphics[width=0.95\linewidth]{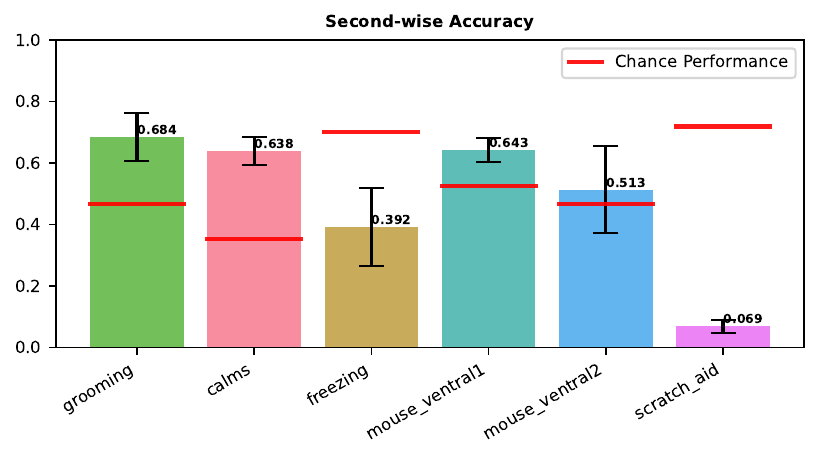}
    \caption{Per second accuracy}
    \label{fig:pro_sa}
\end{figure}

\begin{figure}[!h]
    \centering
    \includegraphics[width=0.95\linewidth]{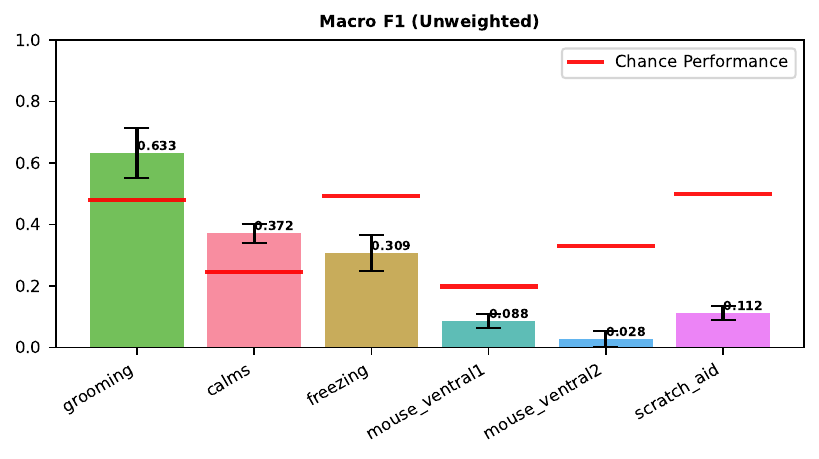}
    \caption{Macro F1 score}
    \label{fig:pro_mf1}
\end{figure}

\newpage

\begin{figure}[!h]
    \centering
    \includegraphics[width=0.95\linewidth]{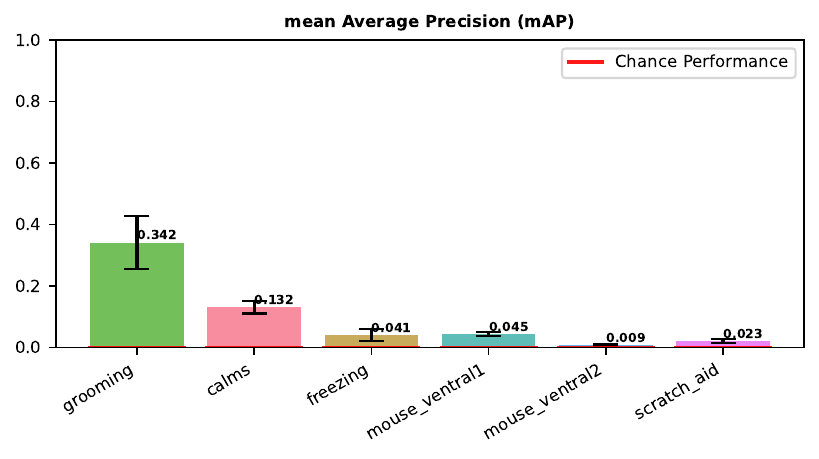}
    \caption{Per second accuracy}
    \label{fig:pro_mAP}
\end{figure}

\begin{figure}[!h]
    \centering
    \includegraphics[width=0.95\linewidth]{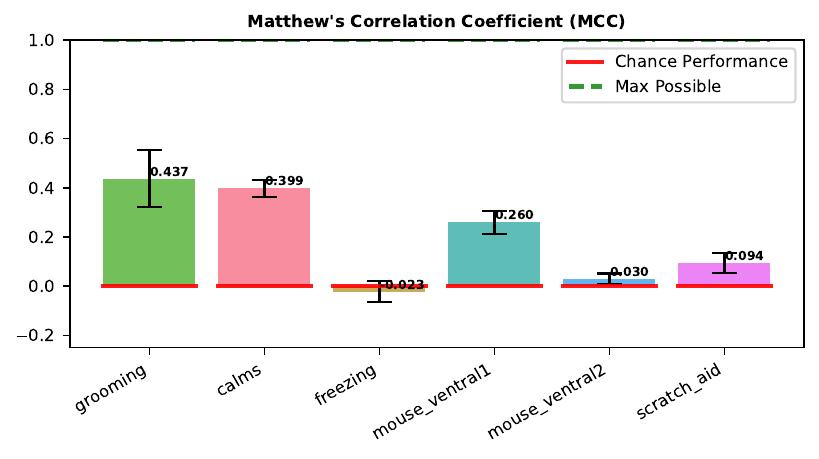}
    \caption{Matthew's Correlation Coefficient}
    \label{fig:pro_mcc}
\end{figure}

\newpage

\begin{figure}[!h]
    \centering
    \includegraphics[width=0.95\linewidth]{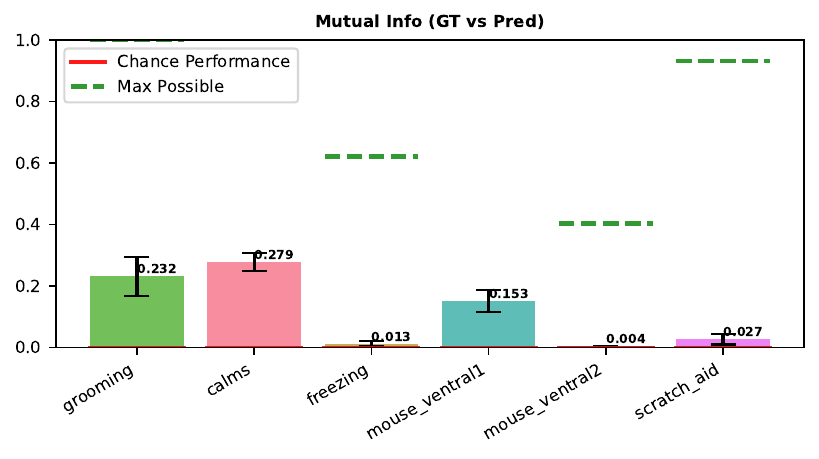}
    \caption{Mutual information between ground truth and predictions}
    \label{fig:pro_MI}
\end{figure}

\newpage

\subsection{Rodent-Bench-Short}

\begin{figure}[!h]
    \centering
    \includegraphics[width=0.95\linewidth]{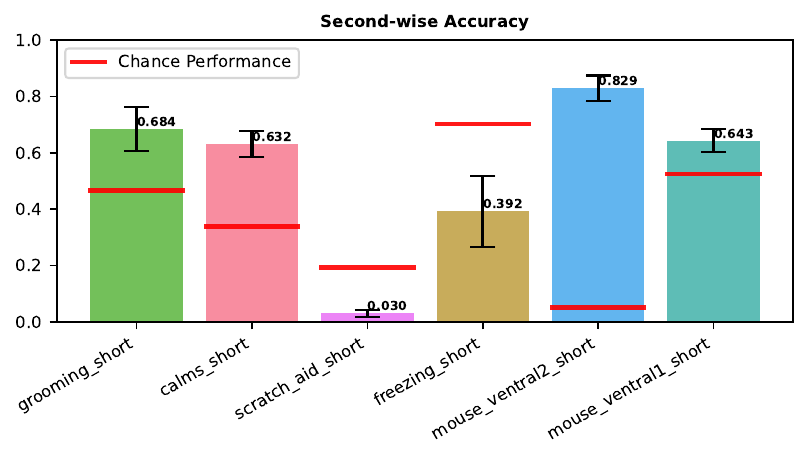}
    \caption{Per second accuracy}
    \label{fig:pro_sa_short}
\end{figure}

\begin{figure}[!h]
    \centering
    \includegraphics[width=0.95\linewidth]{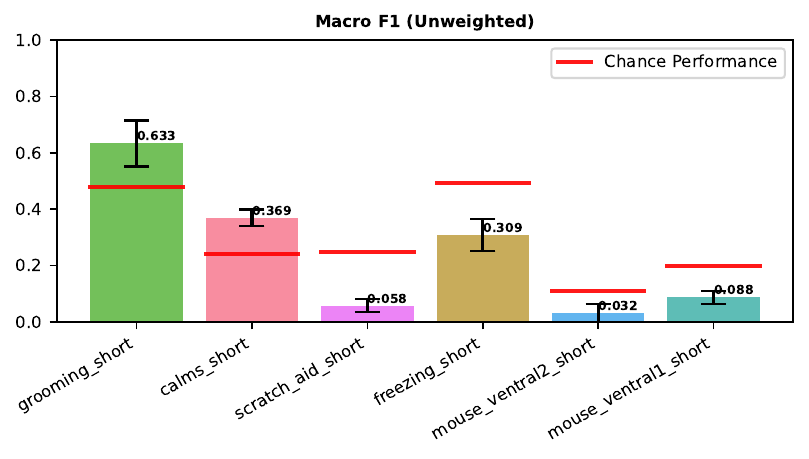}
    \caption{Macro F1 score}
    \label{fig:pro_mf1_short}
\end{figure}

\newpage
\begin{figure}[!h]
    \centering
    \includegraphics[width=0.95\linewidth]{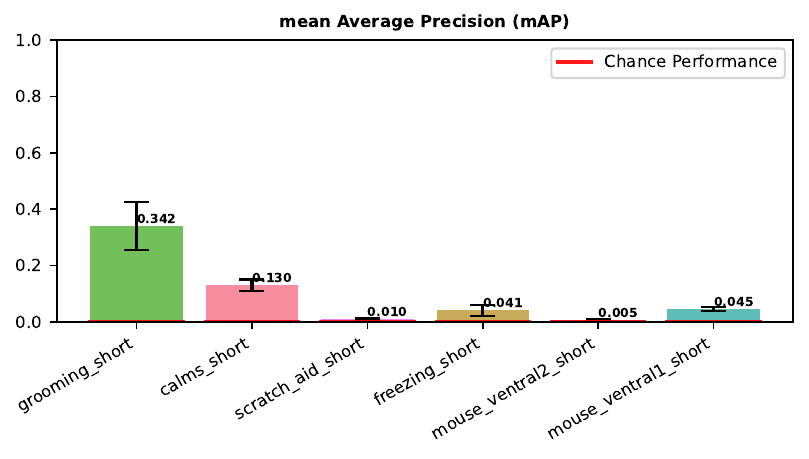}
    \caption{Per second accuracy}
    \label{fig:pro_mAP_short}
\end{figure}

\begin{figure}[!h]
    \centering
    \includegraphics[width=0.95\linewidth]{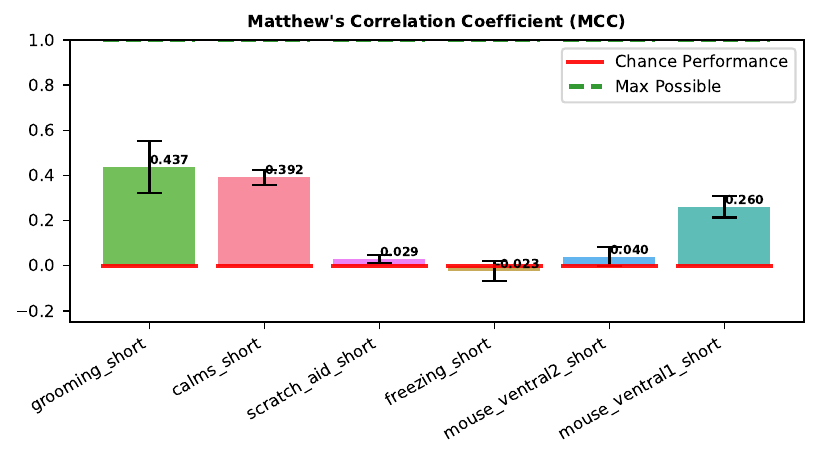}
    \caption{Matthew's Correlation Coefficient}
    \label{fig:pro_mcc_short}
\end{figure}

\newpage

\section{Gemini-Flash Results}
\label{app:flash}

\subsection{Rodent-Bench}

\begin{figure}[!h]
    \centering
    \includegraphics[width=0.95\linewidth]{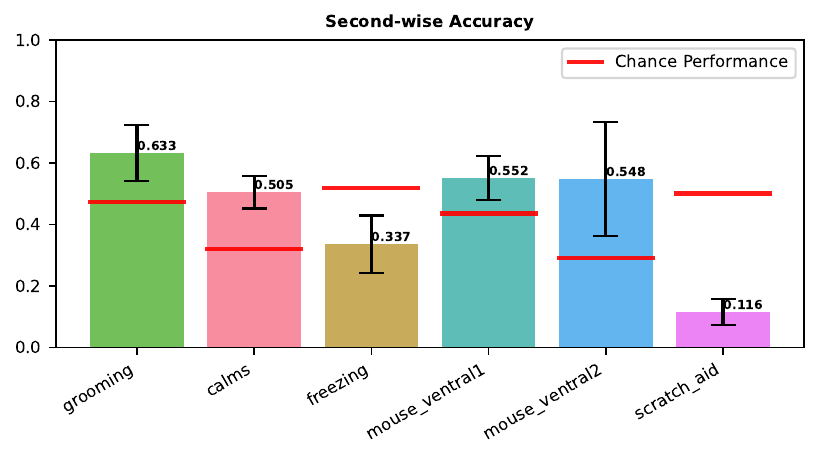}
    \caption{Per second accuracy}
    \label{fig:flash_sa}
\end{figure}

\begin{figure}[!h]
    \centering
    \includegraphics[width=0.95\linewidth]{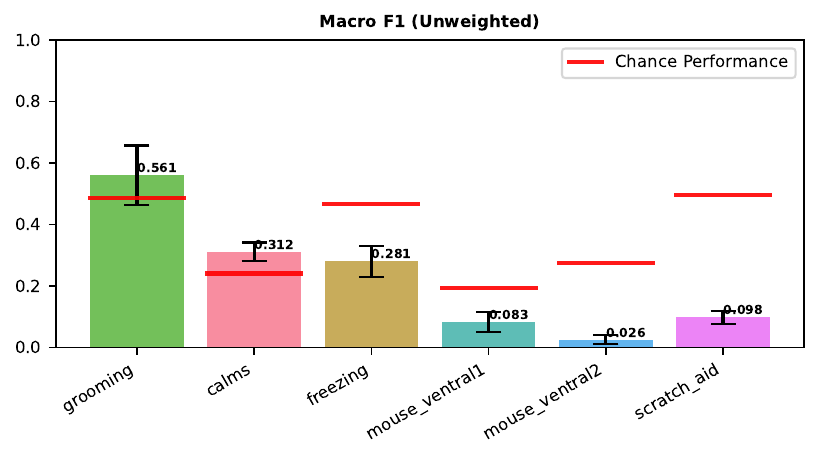}
    \caption{Macro F1 score}
    \label{fig:flash_mf1}
\end{figure}

\newpage

\begin{figure}[!h]
    \centering
    \includegraphics[width=0.95\linewidth]{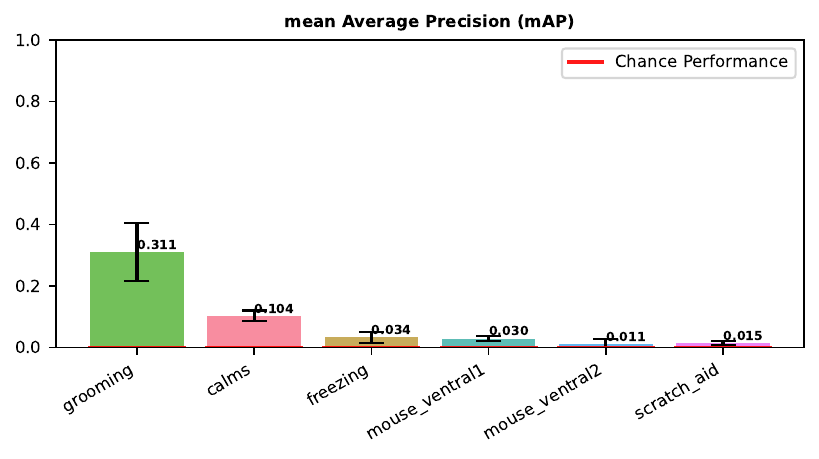}
    \caption{Per second accuracy}
    \label{fig:flash_mAP}
\end{figure}

\begin{figure}[!h]
    \centering
    \includegraphics[width=0.95\linewidth]{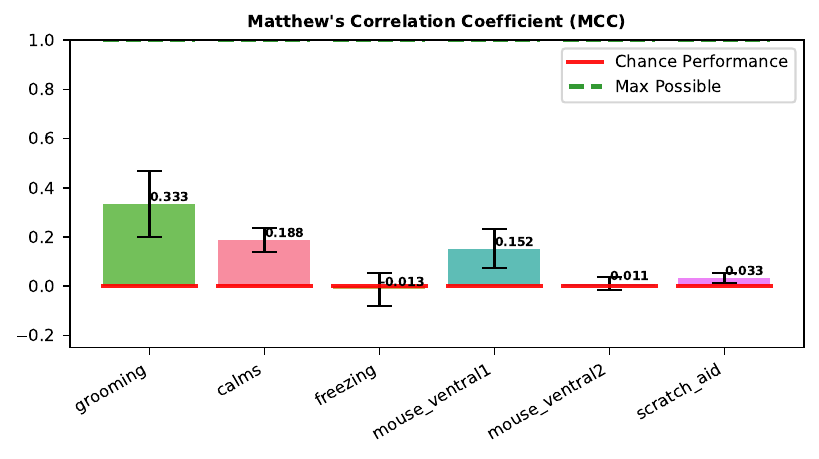}
    \caption{Matthew's Correlation Coefficient}
    \label{fig:flash_mcc}
\end{figure}

\newpage

\begin{figure}[!h]
    \centering
    \includegraphics[width=0.95\linewidth]{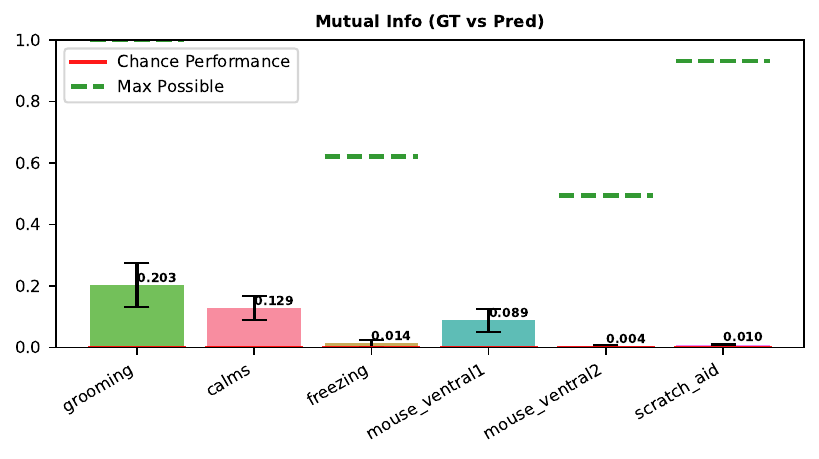}
    \caption{Mutual information between ground truth and predictions}
    \label{fig:flash_MI}
\end{figure}

\newpage

\subsection{Rodent-Bench-Short}

\begin{figure}[!h]
    \centering
    \includegraphics[width=0.95\linewidth]{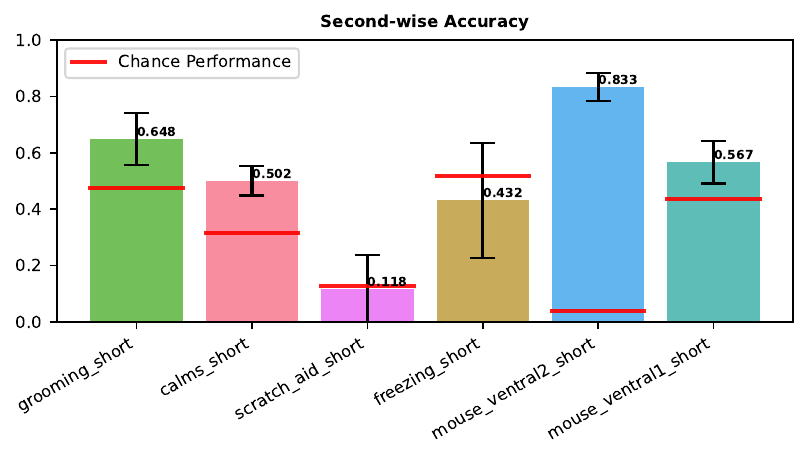}
    \caption{Per second accuracy}
    \label{fig:flash_sa_short}
\end{figure}

\begin{figure}[!h]
    \centering
    \includegraphics[width=0.95\linewidth]{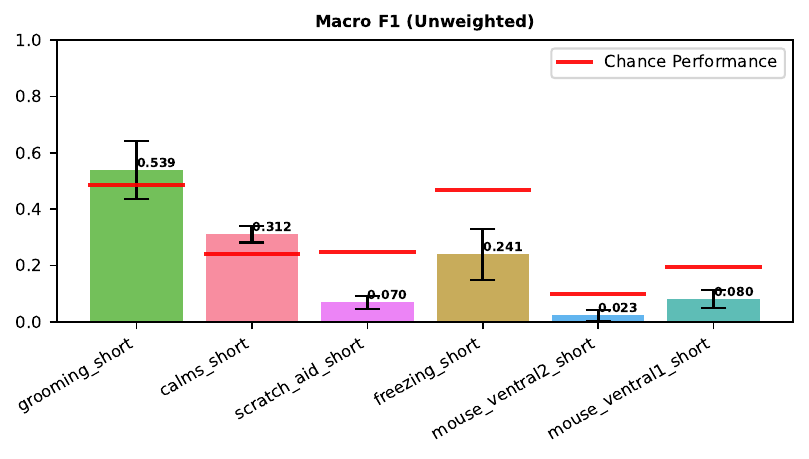}
    \caption{Macro F1 score}
    \label{fig:flash_mf1_short}
\end{figure}

\newpage

\begin{figure}[!h]
    \centering
    \includegraphics[width=0.95\linewidth]{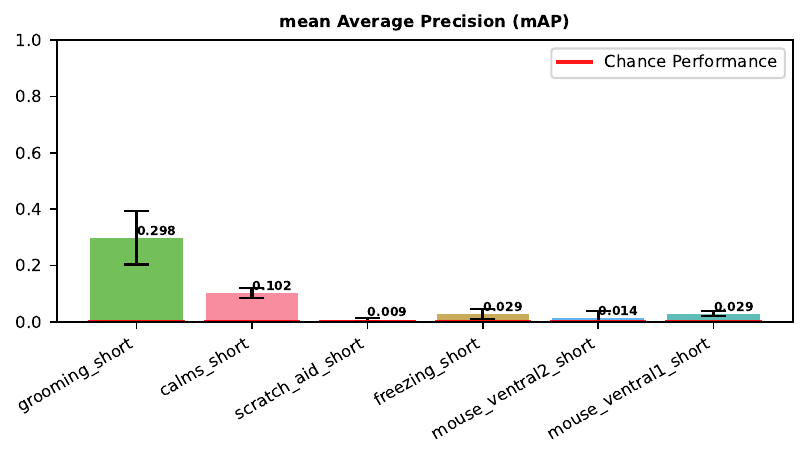}
    \caption{Per second accuracy}
    \label{fig:flash_mAP_short}
\end{figure}

\begin{figure}[!h]
    \centering
    \includegraphics[width=0.95\linewidth]{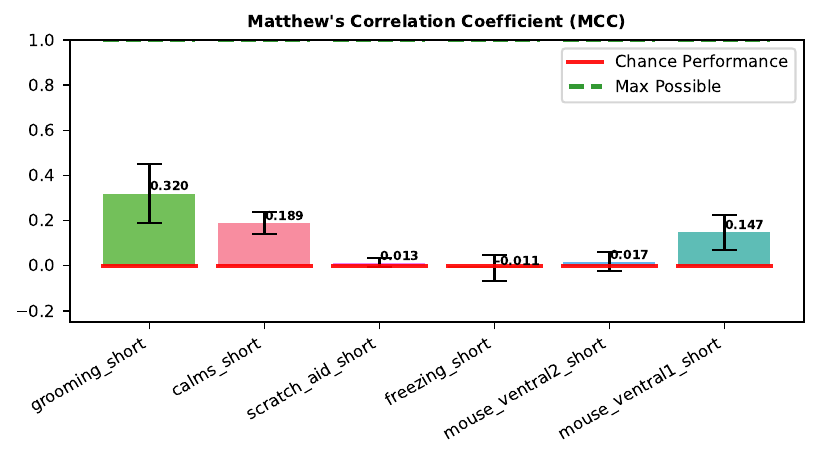}
    \caption{Matthew's Correlation Coefficient}
    \label{fig:flash_mcc_short}
\end{figure}

\newpage

\begin{figure}[!h]
    \centering
    \includegraphics[width=0.95\linewidth]{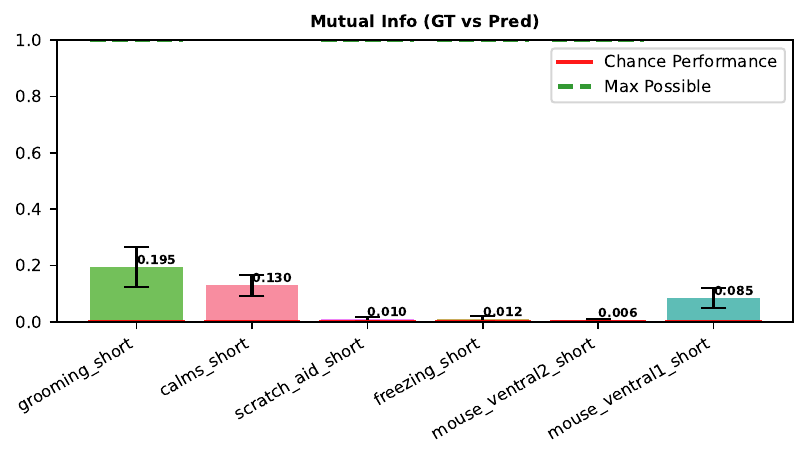}
    \caption{Mutual information between ground truth and predictions}
    \label{fig:flash_MI_short}
\end{figure}

\newpage

\section{Qwen-VL-Max Results}
\label{app:qwen}
Because the Qwen-VL-Max can't ingest videos longer than 10 minutes we only have results for Rodent-Bench-Short.

\subsection{Rodent-Bench-Short}

\begin{figure}[!h]
    \centering
    \includegraphics[width=0.95\linewidth]{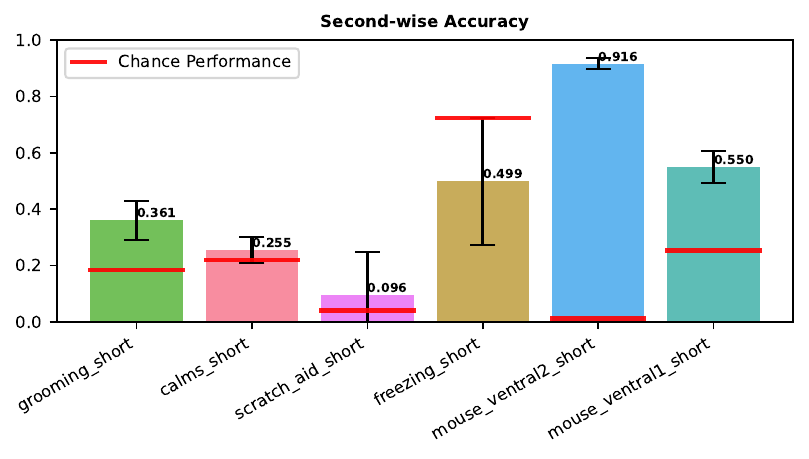}
    \caption{Per second accuracy}
    \label{fig:qwen_sa_short}
\end{figure}

\begin{figure}[!h]
    \centering
    \includegraphics[width=0.95\linewidth]{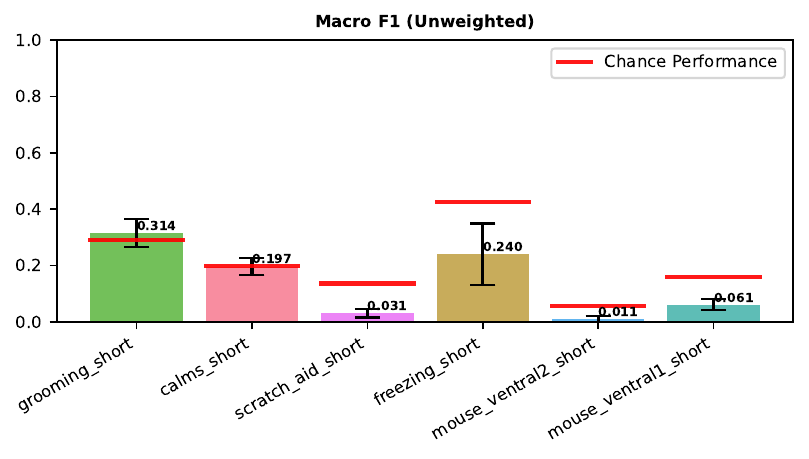}
    \caption{Macro F1 score}
    \label{fig:qwen_mf1_short}
\end{figure}

\newpage

\begin{figure}[!h]
    \centering
    \includegraphics[width=0.95\linewidth]{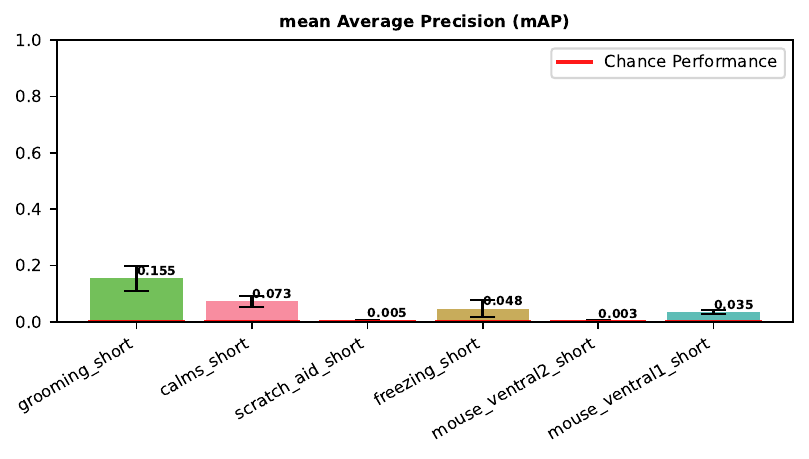}
    \caption{Per second accuracy}
    \label{fig:qwen_mAP_short}
\end{figure}

\begin{figure}[!h]
    \centering
    \includegraphics[width=0.95\linewidth]{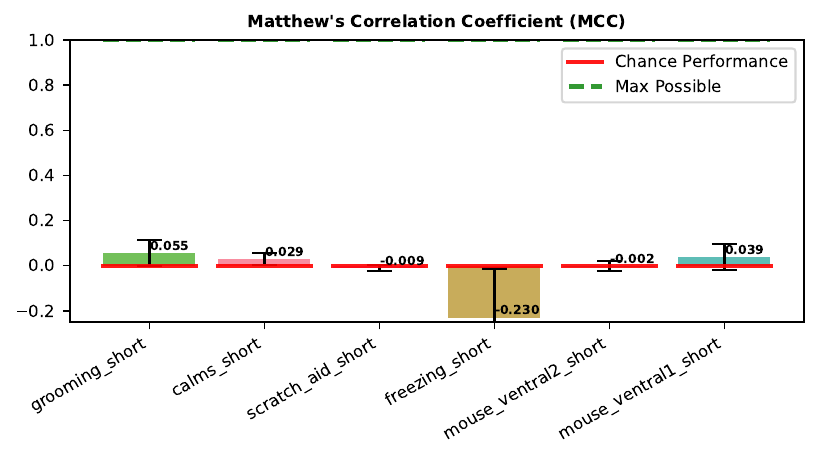}
    \caption{Matthew's Correlation Coefficient}
    \label{fig:qwen_mcc_short}
\end{figure}

\newpage

\begin{figure}[!h]
    \centering
    \includegraphics[width=0.95\linewidth]{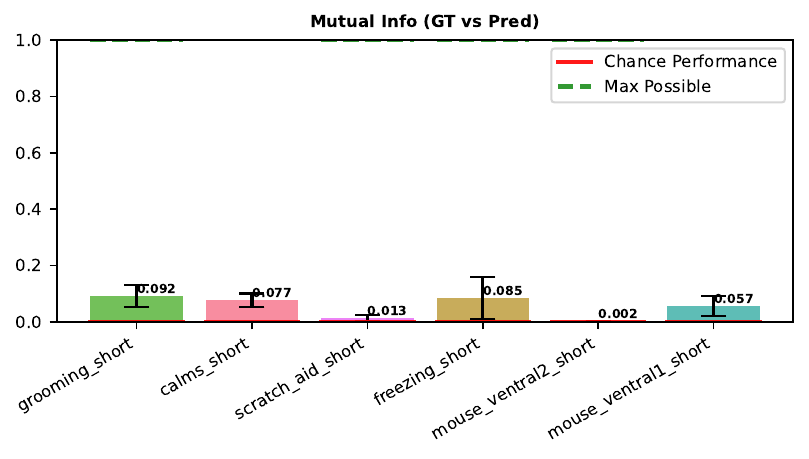}
    \caption{Mutual information between ground truth and predictions}
    \label{fig:qwen_MI_short}
\end{figure}

\newpage

\section{Prompt Templates}
\label{app:prompts}

We provide the complete prompt templates used for each dataset. All prompts follow a consistent structure: role definition, task description, available behavior labels, formatting requirements, and JSON output schema.

\subsection{CalMS21 Social Behaviors}

\begin{lstlisting}[basicstyle=\footnotesize\ttfamily, breaklines=true]
You are a Rodent Behavior Labeler specializing in rodent social behavior.
Your task is to analyze a video of rodents and segment it into periods of distinct behaviors.

Available behavior labels:
- attack - when the black rodent is attacking another rodent
- investigation - when the black rodent is investigating another rodent
- mount - when the black rodent is mounting another rodent
- other - when the black rodent is doing something else

Important: 
You must use ONLY the labels listed above. Do not create new labels or modify existing ones.

Start your analysis from the start of the video and continue until the end of the video.

For each segment, provide:
- segment number (in order)
- start and end time in MM:SS format
- behavior label (must be one of the above labels)

Your response must be in JSON format with the following structure:
{
    "segments": [
        {
            "start_time": MM:SS,
            "end_time": MM:SS,
            "behavior": "behavior_label",
            "segment_number": INTEGER,
        },
        ...
    ]
}
\end{lstlisting}

\subsection{Scratch-AID}

\begin{lstlisting}[basicstyle=\footnotesize\ttfamily, breaklines=true]
You are a Rodent Behavior Labeler specializing in telling when a rodent is scratching.
Your task is to analyze a video of rodents and segment it into periods of distinct behaviors.

Available behavior labels:
- scratching - when the rodent is scratching, usually with the hind legs
- not scratching - when the rodent is not scratching.

Important: 
You must use ONLY the labels listed above. Do not create new labels or modify existing ones. 

Start your analysis from the start of the video and continue until the end of the video. 
The video is of a rodent and is taken from below.

For each segment, provide:
- start and end time in MM:SS format
- segment number (in order)

Your response must be in JSON format with the following structure:
{
    "segments": [
        {
            "segment_number": INTEGER,
            "start_time": MM:SS,
            "end_time": MM:SS,
            "behavior": "behavior_label",
        },
        ...
    ]
}
\end{lstlisting}

\subsection{Rodent Grooming Detection}

\begin{lstlisting}[basicstyle=\footnotesize\ttfamily, breaklines=true]
You are a Rodent Behavior Labeler specializing in identifying grooming behaviors in rodents.
Your task is to analyze a video of rodents and segment it into periods of distinct behaviors.
The video shows a rodent from above.

Available behavior labels:
- grooming - when the rodent is actively grooming itself (e.g., licking, scratching, cleaning fur)
- other - when the rodent is not grooming (e.g., walking, exploring, resting)

Important: 
You must use ONLY the labels listed above. Do not create new labels or modify existing ones.
Grooming behaviors are characterized by:
- Repetitive movements of paws over the face or body
- Licking of fur or paws
- Scratching with hind legs
- Cleaning of specific body parts

Start your analysis from the start of the video and continue until the end of the video.

For each segment, provide:
- segment number (in order)
- start and end time in MM:SS format
- behavior label (must be one of the above labels)

Your response must be in JSON format with the following structure:
{
    "segments": [
        {
            "segment_number": INTEGER,
            "start_time": MM:SS,
            "end_time": MM:SS,
            "behavior": "behavior_label",
        },
        ...
    ]
}
\end{lstlisting}

\subsection{Freezing Behavior}

\begin{lstlisting}[basicstyle=\footnotesize\ttfamily, breaklines=true]
You are a Rodent Behavior Labeler specializing in identifying freezing behaviors in rodents.
Your task is to analyze a video of rodents and segment it into periods of distinct behaviors.
The video shows a rodent from above.

Available behavior labels:
- Freezing - when the rodent is Freezing, i.e characterized by the complete cessation of movement, except for respiratory-related movements so no head twitching for instance.
- Not Freezing - when the rodent is not Freezing

Important: 
You must use ONLY the labels listed above. Do not create new labels or modify existing ones.

Start your analysis from the start of the video and continue until the end of the video.

For each segment, provide:
- segment number (in order)
- start and end time in MM:SS format
- behavior label (must be one of the above labels)

Your response must be in JSON format with the following structure:
{
    "segments": [
        {
            "segment_number": INTEGER,
            "start_time": MM:SS,
            "end_time": MM:SS,
            "behavior": "behavior_label",
        },
        ...
    ]
}
\end{lstlisting}





\end{document}